\pdfoutput=1

\documentclass[11pt]{article}

\usepackage[]{acl}

\usepackage{times}
\usepackage{latexsym}
\usepackage{tabularx}
 \usepackage{booktabs} 
 \usepackage{enumitem}

 \usepackage{float}

\usepackage[T1]{fontenc}

\usepackage[utf8]{inputenc}

\usepackage{microtype}

\usepackage{inconsolata}

\usepackage{graphicx}

\usepackage{xspace}

\definecolor{primary3}{HTML}{D38547}
\definecolor{complementary3}{HTML}{2B807D}
\definecolor{both3}{HTML}{973384}

\definecolor{sigbar}{HTML}{de8f9c}

\newcommand{\dialogue}{\textcolor{primary3}{dialogue}\xspace}
\newcommand{\dialogues}{\textcolor{primary3}{dialogues}\xspace}
\newcommand{\Dialogue}{\textcolor{primary3}{Dialogue}\xspace}
\newcommand{\dialog}{\textcolor{complementary3}{dialog}\xspace}
\newcommand{\Dialog}{\textcolor{complementary3}{Dialog}\xspace}
\newcommand{\both}{\textcolor{both3}{both}\xspace}
\newcommand{\Both}{\textcolor{both3}{Both}\xspace}

\newcommand{\dialogns}{\textcolor{complementary3}{dialog}}
\newcommand{\dialoguens}{\textcolor{primary3}{dialogue}}

\newcommand{\etc}{etc.}
\newcommand{\StandardStartYear}{2010\xspace}
\newcommand{\NumPapersTotal}{87,498\xspace}
\newcommand{\NumPapersCs}{52,249\xspace}
\newcommand{\AverageNumAuthors}{4.6\xspace}
\newcommand{\DialogCorpusSize}{7656\xspace}
\newcommand{\NumTopVenues}{25\xspace}
\newcommand{\PercentPubDialogue}{72\%\xspace}
\newcommand{\PercentPubBoth}{5\%\xspace}
\newcommand{\PercentPubDialog}{24\%\xspace}
\newcommand{\TopAuthorsPapeThreshold}{20\xspace}
\newcommand{\NumTopAuthors}{15\xspace}
\newcommand{\TopAuthorsDialogFrac}{0.248\xspace}

\newcommand{\AmericanLessThanChina}{12.9\xspace}
\newcommand{\AmericanLessThanChinaCiLow}{19.7\xspace}
\newcommand{\AmericanLessThanChinaCiHigh}{6.1\xspace}
\newcommand{\AmericanLessThanGb}{23.0\xspace}
\newcommand{\AmericanLessThanGbCiLow}{33.3\xspace}
\newcommand{\AmericanLessThanGbCiHigh}{10.9\xspace}
\newcommand{\NumWordcombos}{38\xspace}

\newcommand{\NumDialogWordcombos}{31\xspace}

\newcommand{\DialogWordcomboFrac}{81.6\xspace}
\newcommand{\DialogWordcomboFracCiLow}{68.4\xspace}
\newcommand{\DialogWordcomboFracCiHigh}{92.1\xspace}

\title{``\Dialogue'' vs ``\Dialog'' in NLP and AI research: \\ Statistics from a Confused Discourse}

\author{David Gros \\
  University of California, Davis \\
  \texttt{dgros@ucdavis.edu}
  }

\begin{document}
\maketitle
\begin{abstract}
Within computing research, there are two spellings for an increasingly important term – \dialogue and \dialog. We analyze thousands of research papers to understand this ``dialog(ue) debacle''. Among publications in top venues that use ``dialog(ue)'' in the title or abstract, \PercentPubDialogue use ``\dialogue'', \PercentPubDialog use ``\dialog'', and \PercentPubBoth use \Both in the same title and abstract. This split distribution is more common in Computing than any other academic discipline. We investigate trends over \textasciitilde 20 years of NLP/AI research, not finding clear evidence of a shift over time. Author nationality is weakly correlated with spelling choice, but far from explains the mixed use. Many prolific authors publish papers with both spellings. 
We use several methods (such as syntactic parses and LM embeddings) to study how dialog(ue) context influences spelling, finding limited influence.
Combining these results together, we discuss different theories that might explain the dialog(ue) divergence.

\end{abstract}

\section{Introduction}\label{sec:intro}

Computer scientists have a bit of a silent spelling spat. It has become more apparent as systems that let people talk to computers are going mainstream (starting with Apple Siri, Amazon Alexa, \etc – now with ChatGPT, Claude, \etc). Yet the people making these systems can not seem to decide if these are \dialogue systems or \dialog systems.

\begin{figure}[t]
    \centering
    \includegraphics[width=0.99\columnwidth]{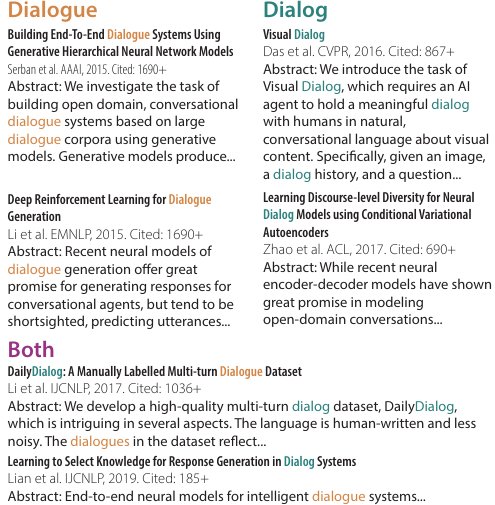}
    \caption{Examples of varying uses of dialog(ue) in prominent NLP/AI research.}
    \label{fig:dialogexamples}
    \vspace{-1.5em}
\end{figure}

\newcommand{\paper}[6]{%
    {\small 
    \noindent\textbf{#1} 
    \textit{#2} 
    \textit{#3, #4.} 
    Cited: #5+. 
    \textbf{Abstract:} #6\\[2ex] 
    } 
}


Looking at top AI and NLP conferences we see prominent papers on each side (\autoref{fig:dialogexamples}). More surprisingly, even just considering the title and abstract of a paper, some prominent papers use \both!

Thus, when writing prose or code, it is unclear which one to choose. When reviewing dictionaries \cite{MerriamWebster2024, oxford2023dialogue}, web grammar guides \cite{LanguageTool2023, writingexplained2024dialog}, and online discussion forums, we see roughly three explanations:

\newcounter{ExplanationCounter}

\newcommand{\Explanation}[2]{%
    \refstepcounter{ExplanationCounter} 
    \texttt{Explanation-\arabic{ExplanationCounter}} \textit{#1:} \label{#2}
}

\newcommand{\refexplan}[1]{\texttt{Explanation-\ref{#1}}}

\Explanation{American English}{exp:american_english}
Like “catalogue” and “catalog”, some suggest it is a difference between British and American English. Yet, this is unclear. The American Merriam-Webster Dictionary lists \dialog as a “less common variant”
. The Oxford English Dictionary lists both spellings.

\Explanation{Computing Specific}{exp:computing_specific}
It's commonly said “\dialog box” is the right spelling for the popup boxes in applications. Some guides generalize that \dialog is preferred in all computing.

The exact entomology here is unclear. "\Dialog box" is used in early GUI systems, such as the manual for the Apple Lisa \cite{apple1983lisa}, and in manuals for Windows 1.0 \cite{windowssdk}. The term does not appear in the manual for the Xerox Star \cite{xerox1981star}. The 1979 manual for the Xerox Alto refers to a "\dialogue'', but as a text prompt \cite{shenguiding, xerox1979alto}. Thus, it seems plausible that "\dialog box", no "ue" included, is a neologism created by Apple engineers in the mid-1980s that persisted.

\Explanation{Completely Interchangeable}{exp:interchangeable}
Most sources suggest the spellings are interchangeable and does not matter. This seems mostly true. Yet as the discourse moves from research papers into actual code implementing these systems, the selection \textit{does} matter. In source code, intermixing the spellings can cause bugs.

\vspace{0.25em}

We seek to understand if these explanations are valid. Additionally we consider some linguistic theories for language change. For example, that languages change for improved ``economy'' when pushing for easier communication, or ``analogy'' when trying to simplify a language's regularity \cite{Deutscher2005TheUO}. Orthography is tied to many social influences \cite{sebba2007spelling}.

Guided by the three \texttt{Explanations} above and linguistic theory, we study four research questions:

\newcounter{RQCounter}

\newcommand{\RQ}[2]{%
\refstepcounter{RQCounter} \label{#1}
	\vspace{0.02in} \noindent \textbf{RQ\arabic{RQCounter}.~#2} 
}

\RQ{rq:diffs}{What is the distribution of dialog(ue) orthography in NLP and AI research?} \S\ref{sec:aggstats}

\RQ{rq:shift}{Is there an ongoing othography shift over time?} \S\ref{sec:shift}

\RQ{rq:author}{Is there author-level influences (such as an American English choice) that explains the differences?} \S\ref{sec:authortrend}

\RQ{rq:semantics}{Are the spellings truely interchangeable, or is each preferred in certain contexts?} \S\ref{sec:context}


\paragraph{Contributions} This study was motivated by a combination of intellectual curiosity of a linguistic quirk in NLP/AI, as well as some first-hand frustration when deciding which spelling to choose in prose and code. Our contributions following the above RQs help inform answers. In addition, our work has auxiliary contributions like giving a quantitative overview of dialog(ue)-specific literature (along dimensions like Venue, Year, Author, \etc), and a framework for further meta-study of dialog(ue) research. Additionally, our methods for comparing contextual influence on spelling might be informative for similar computational linguistic work.

\section{Methodology}

We start with some definitions. An academic work with \dialogns(s) or \dialoguens(s) in the title or abstract is defined as a “Dialog(ue) Paper”. We use the Semantic Scholar (S2) search API \cite{Kinney2023TheSS} to retrieve “Dialog(ue) Papers”. The S2 Corpus assigns the paper to one or more fields of study, and a publication venue. Data is retrieved in March 2024. In total, \NumPapersTotal works are retrieved, with \NumPapersCs having the field “Computer Science” (CS).

Our focus is the use of dialog(ue) within NLP and AI research. 
We use a process to identify ``High Impact Dialog(ue) Venues" and corresponding peer-reviewed CS ``Dialog(ue) Publictions'' (methods in \autoref{sec:filtvenue}). With the exception of \autoref{sec:shift}, we filter to papers after \StandardStartYear.

\begin{table}[htb]
\tiny
\centering
\begin{tabular}{llrr}
\toprule
& Venue & Count & Avg. Citations\\
\midrule
1 & ACL & 615 & 58.7\\
2 & EMNLP & 508 & 39.7\\
3 & SIGDIAL & 296 & 30.8\\
4 & AAAI & 267 & 45.5\\
5 & NAACL & 231 & 41.8\\
... & ... & ... & ...\\
21 & CoNLL & 18 & 40.4\\
22 & TSLP & 15 & 25.3\\
23 & ICML & 14 & 46.3\\
24 & LAK & 13 & 80.6\\
25 & CHIIR & 11 & 26.4\\
\bottomrule
\end{tabular}
\caption{A subset of the High Impact Dialog(ue) Venues. Data starts in 2010. See appendix for full table.}\label{tab:topvenues}
\vspace{-2em}
\end{table}

\section{Aggregate Statistics}\label{sec:aggstats}

\begin{figure}[h!]
    \centering
    \includegraphics[width=0.99\columnwidth]{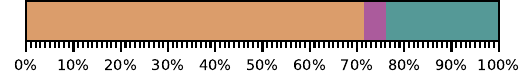}
    \caption{Distribution for CS Dialogu(ue) Publications.}
    \label{fig:aggfig}
\end{figure}

We see that \dialogue is dominant in Dialog(ue) Publications, representing \PercentPubDialogue of the publications. \Both represents a surprising \PercentPubBoth . \Dialog is about a quarter of the publications.

We can compare Computing to other disciplines. We do not have selected venues for all disciplines, so we instead chose works with a citation count in the top 75th-percentile of each field. We consider the first field returned by S2. 

\begin{figure}[tbh]
    \centering
    \includegraphics[width=0.99\columnwidth]{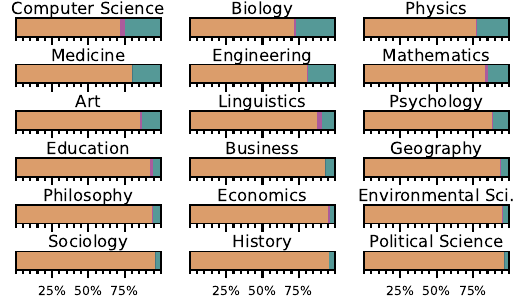}
    \caption{Dialog(ue) Papers across disciplines.}
    \label{fig:disciplines}
    \vspace{-0.5em}
\end{figure}

Computing has the lowest \dialogue use, and a much larger use of \both. One might think this gives evidence towards \refexplan{exp:computing_specific}, but the similar use in fields like Biology diminishes such conclusions. For the rest of the paper we focus on AI/NLP research.

\section{A Search for a Shift}\label{sec:shift}

Languages can evolve over time.
We look to see if there is an ongoing orthography shift within the computing research community.

\begin{figure}[tb]
    \centering
    \includegraphics[width=0.99\columnwidth]{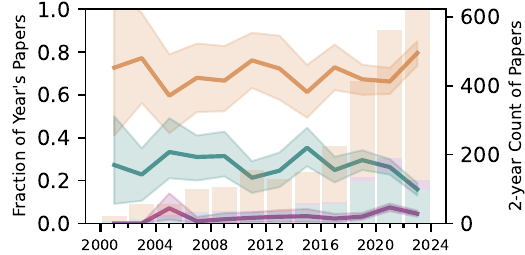}
    \caption{CS Dialog(ue) Publications by Year. To reduce noise, we group into 2-year intervals. Shaded line area is a 95\% 2-year bootstraped CI.}
    \label{fig:year}
    \vspace{-1em}
\end{figure}

\autoref{fig:year} shows exclusive
\dialogue might have reached its lowpoint around $[2014, 2016)$ and has since increased to \textasciitilde 77\% in $[2022, 2024)$.
Exclusive \dialog use has appeared to trend downwards, with \both use
making up a larger fraction.
There were fewer than 100 dialog(ue) publications per year before 2017, leading to high uncertainty of older fractions. Thus, with current evidence we do not conclude there is an orthography shift towards \dialog over the last 24 years of NLP/AI research. If heavy usage of dialog(ue) continues, it might be clearer if there is actually a shift. We note how this data might differ from general usage in books \cite{googlengrams}, where \dialog use fraction peaked at around 40\% in 2005, and rapidly declined to 19\% in 2019 (see \autoref{append:ngrams}).

\section{Trends by Author}\label{sec:authortrend}

\subsection{Individual Author Consistency}

\begin{figure}[tbh]
    \centering
    \includegraphics[width=0.99\columnwidth]{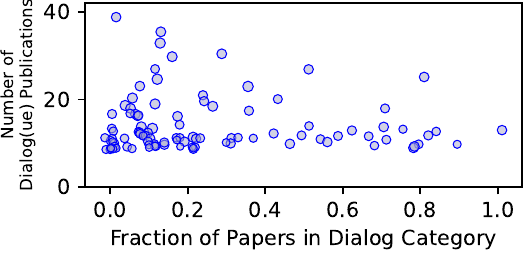}
    \caption{Distribution for 100 authors with most CS Dialog(ue) Publications.
    }
    \label{fig:authorteam}
    \vspace{-1em}
\end{figure}

For each researcher we calculate the fraction of their Dialog(ue) Publications in the \dialog{} category, with papers in the \Both{} category counting as 0.5 in the average.\footnote{Note, due to misidentification in the S2 corpus, some authors have their works split between multiple authorids.}

Most authors with many Dialog(ue) Publications have used a mix of forms. Among the \NumTopAuthors authors with at least \TopAuthorsPapeThreshold papers, the mean author \Dialog fraction is \TopAuthorsDialogFrac, similar to the overall average.

We note that the average Dialog(ue) Publication has \AverageNumAuthors authors. In Computing / the Sciences, coauthorship is typically broad. Thus, the results here are not necessarily reflective of a researcher's individual choice, but of a rough aggregate of their network of collaborators.
Still, we find the intermixing of spelling to be interesting.

\subsection{Nationality Influence}\label{natinfluence}

Next we explore the \refexplan{exp:american_english} that use of \dialog{} is influenced by an American English preference (i.e., a "catalog" vs "catalogue" factor).

We download PDFs from open access Dialog(ue) Publications, process the text, and estimate author institution information using GPT-3.5 (more details in \autoref{append:nationalityinstitute}).

\begin{figure}[tbh]
    \centering
    \includegraphics[width=0.99\columnwidth]{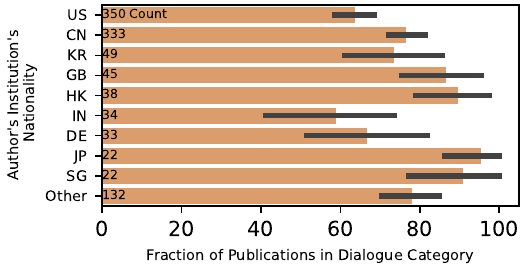}
    \caption{Author nationality with a 95\% CI on \dialogue}
    \label{fig:nationality}
\end{figure}

We find some differences from author institution. Authors at American institutions use \dialogue \AmericanLessThanChina percent-points less than authors at Chinese institutions (diff C95 \AmericanLessThanChinaCiHigh --\AmericanLessThanChinaCiLow), and \AmericanLessThanGb percent-points less than authors at British intuitions (diff C95 \AmericanLessThanGbCiHigh --\AmericanLessThanGbCiLow). However, author institution country has limited overall predictive power\footnote{logistic prediction of \dialogue has 0.037 $R^2_{McFadden}$. Though LLR $p < 0.001$ , confirming some slight signal.}.
    
It might be that author's spelling choice is influenced not by their current institution, but where they learned English spelling. We use LLMs with web-search to attempt to estimate author home nationality. However, we find identification to be noisy, and counts have low statistical power. If there is an influence, it is likely small. (\autoref{append:nationality})

Thus, there is some evidence of an "American English influence", but it is fairly weak, and does not explain the dialog(ue) debate.

\section{Does Context Influence Spelling?}\label{sec:context}

Next, we consider different contexts / word senses of dialog(ue) to see if it influences spelling, testing \refexplan{exp:interchangeable} that the spellings carry no meaning and are interchangeable.

To do this, we build a "dialog(ue) corpus", which are uses of "dialog(ues)" and context. From the Dialog(ue) Publications, we gather \DialogCorpusSize uses from the titles/abstracts.

\subsection{Noun Phrase Use}

\begin{figure}[tbh]
    \centering
    \includegraphics[width=0.99\columnwidth]{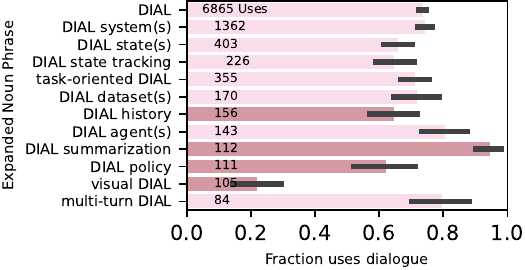}
    \caption{Noun phrases with a 95\% CI. {\textcolor{sigbar}{Darker shaded }}values statistically significant. See appendix for more.}
    \label{fig:phrase}
\end{figure}

We use the Spacy dependency parser \cite{spacy3} to extract noun phrases\footnote{specifically, "noun chunks" in Spacy}. A noun phrase is a noun with additional words describing the noun (for example,
"conversational \dialog system" or "friendly \dialogue").
Because some noun phrases can be highly specific (eg, "the open-domain Korean \dialogue{} model"), we analyze sub-noun-chunks considering all subsets of neighboring words containing dialog(ue). We filter to sub-chunks used by at least 50 authors. See \autoref{append:nounphrase} for more details.

Generally the phrase context of dialog(ue) is not influential on spelling. However, a logistic model identifies 4 significant\footnote{$\alpha=0.05$ after applying Benjamini-Hochberg correction \cite{Benjamini1995ControllingTF}} phrases (Appendix \ref{append:nounlinear}). The most significant phrase, "* visual dialog(ues) *", calls back to the most cited \dialog work in \autoref{fig:dialogexamples}.


The phrase "\dialog{} box" does not appear in the corpus, thus limiting \refexplan{exp:computing_specific}.

\subsection{LM Contextual Prediction}

For each use in our Dialog(ue) Corpus, we mask out dialog(ue) tokens, and get contextual embeddings with RoBERTa \cite{Liu2019RoBERTaAR}. We fit a classifier using these embeddings. See \autoref{append:lm} for methodology. The model gives no improvement over the base rate, suggesting context has limited influence on dialog(ue) spelling.

\subsection{Location and Morphology}
We find no significant difference on location of use (ie, title or abstract). We do find that plural use is more likely to be \dialogues. Proper nouns are more likely to be \Dialog. Additionally, closed compound words (eg, DailyDialog, MedDialog, etc) are more likely to \dialog. See \autoref{append:morphology} for more.

\subsection{Use In Source Code}

Spelling is important when writing software in order to avoid bugs. We analyze source code corpora, finding an increased preference for \dialog compared to NLP/AI paper titles/abstracts. See \autoref{sec:sourcecode}.

\section{Related Work}

In \autoref{sec:intro} we review several sources on \dialog and \dialogue. Due to space, further review is moved to appendix.



We are not aware of any similar formal and methodical scholarship on "dialog(ue)" orthography.

\section{Discussion and Conclusion}

Our study highlights the surprising amount of disagreement on a core word for the field. The reasons for this appear to not fit neatly into one explanation.

From the prose uses, we not do find evidence \refexplan{exp:computing_specific} is a direct influence given lack of consistent noun-phrases like ``\dialog box'' (or other mined phrases), and carryover to fields like Biology and Physics. We also do not find evidence that the distribution is due to a spelling shift happening in the middle.

The same-author mixing suggests a lack of awareness or indifference on the selection of the spelling. We find \dialog use is partially due to American English influences. Given evidence from source code and word compounds, we speculate computer scientists might be particularly motivated by economy while code switching between NL/PL.

Though the spelling mixing is surprising, it is clear the community is accepting of both spellings. In this work we thoroughly explore this phenomena and give a descriptive guide, but refrain from trying to argue which spelling is {\textcolor{complementary3}{bett}{\textcolor{primary3}{er}. We hope this improves understanding of an interesting linguistic phenomena happening inside our community.



\bibliography{acl_latex}

\appendix

\section*{Appendix}\label{sec:appendix}

\section{Filter by venues}\label{sec:filtvenue}

We try to focus on major NLP/AI research publications. We first filter to CS focused venues (85\% of the first field is CS), and with at least 10 Dialog(ue) Papers. We then calculate the average number of citations of Dialog(ue) Papers grouped by venue.  We take the top \NumTopVenues venues by mean Dialog(ue) Paper citations. These venues are labeled as “High Impact Dialog(ue) Venues”. This criteria admits some non-NLP/AI venues (such as CHI, and The Web Conference), but this represents a smaller fraction of papers. The venues are listed in \autoref{tab:topvenuesfull}. A “Dialog(ue) Paper” is referred to as “Top CS Dialog(ue) Publication” (for brevity, a "Dialog(ue) Publication") the venue is one of these venues.


\begin{table*}[htb]
\small
\centering
\begin{tabular}{llrr}
\toprule
& Venue & Papers & Avg. Citations\\
\midrule
1 & Annual Meeting of the Association for Computational Linguistics & 615 & 58.7\\
2 & Conference on Empirical Methods in Natural Language Processing & 508 & 39.7\\
3 & SIGDIAL Conference & 296 & 30.8\\
4 & AAAI Conference on Artificial Intelligence & 267 & 45.5\\
5 & North American Chapter of the Association for Computational Linguistics & 231 & 41.8\\
6 & International Joint Conference on Artificial Intelligence & 161 & 35.7\\
7 & International Conference on Human Factors in Computing Systems & 159 & 60.3\\
8 & Conference of the European Chapter of the Association for Computational Linguistics & 129 & 37.8\\
9 & International Conference on Computational Logic & 64 & 114.8\\
10 & The Web Conference & 60 & 33.0\\
11 & Computer Speech and Language & 56 & 49.5\\
12 & Automatic Speech Recognition \& Understanding & 47 & 23.8\\
13 & Neural Information Processing Systems & 40 & 134.3\\
14 & International Joint Conference on Natural Language Processing & 39 & 42.6\\
15 & International Conference on Learning Representations & 33 & 141.9\\
16 & Dialogue and Discourse & 28 & 44.5\\
17 & Transactions of the Association for Computational Linguistics & 27 & 73.6\\
18 & Knowledge Discovery and Data Mining & 25 & 23.6\\
19 & International Workshop on Semantic Evaluation & 21 & 90.0\\
20 & Computer Vision and Pattern Recognition & 21 & 117.1\\
21 & Conference on Computational Natural Language Learning & 18 & 40.4\\
22 & TSLP & 15 & 25.3\\
23 & International Conference on Machine Learning & 14 & 46.3\\
24 & International Conference on Learning Analytics and Knowledge & 13 & 80.6\\
25 & Conference on Human Information Interaction and Retrieval & 11 & 26.4\\
\bottomrule
\end{tabular}
\caption{High Impact Dialog(ue) venues. We emphasize how paper counts are for our definition of ``Dialog(ue) Papers'' which is when ``dialog(ues)'' is mentioned in the title or abstract. Other papers which use the term in the body, or use synonyms are not included. The average citations from citations of these Dialog(ue) Publications by any later academic work (as identified in the S2 corpus).}
\label{tab:topvenuesfull}
\end{table*}



\section{Books data comparison}\label{append:ngrams}

Using the Google Ngrams Viewer \cite{googlengrams, Lin2012SyntacticAF} we can compare NLP/AI reserach against general usage in a books corpus. This is shown in \autoref{fig:ngrams}. In book data there is a much clearer trend of \dialog use increasing starting in the 1980s but has since rapidly declined. The data from NLP/AI research is less clear. In part this is due to the fact that only in the last 10 years has a substantial number of annual dialog(ue) publications been available.

\begin{figure*}[tbh]
    \centering
    \includegraphics[width=2\columnwidth]{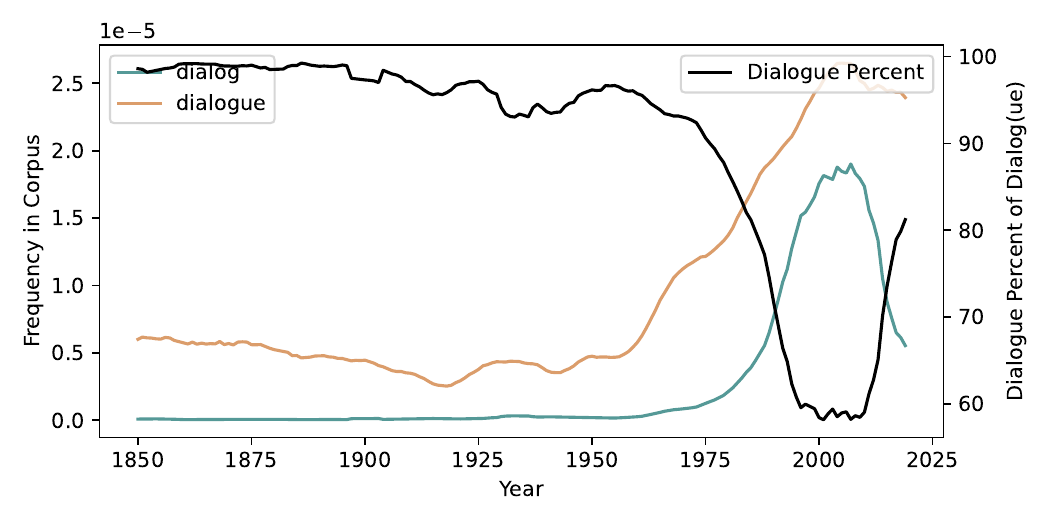}
    \caption{Data from \cite{googlengrams}. This shows data from books indexed by Google.}
    \label{fig:ngrams}
\end{figure*}

\section{Nationality Data Details}\label{append:nationalityinstitute}

We download PDFs from open access Dialog(ue) Publications. Not all PDFs are downloadable (due to many limitations including closed-access publications, some web hosts blocking scraping downloads (despite rate limits), removed or restricted links, etc). Thus, this is a different set of publications than our original set, but the aggregate proportions of \dialogue{}/\dialog{}/\both{} was similar.

For each PDF we parse the text with PyMuPDF \cite{pymupdf}. We then extract author institution and estimated institution country by providing GPT-3.5 with the start of the paper text, and a prompt to extract author information into structured form. We then estimate a country for the paper by a plurality count of each author's institution country, weighting first-author 4x, and last author 2x.

\section{Additional Morphology Info}\label{append:morphology}

\subsection{Location}

We compare occurrences between the title and the abstract. We do not find evidence of a difference (Fisher's Exact Test $p$ > 0.05). A limitation of our study is we do not parse into the bodies of papers\footnote{this is due to a mix of reasons. One is the complexity of handling PDF parses (ideally making sure to eliminate non-author-uses like citations to other work or venue names) that put it out of scope. Additionally, one must deal with the lower coverage of open access pdfs. Focusing on titles and abstracts is also attractive for our RQ3, as it gives higher likelihood most authors of a paper will have visibility to the occurrence if it is in the title and abstract. If needed, these challenges could be addressed.}. Future work could explore body usages. The location serves as a weak proxy for the formality of the writing (with titles most formal and reviewed, followed by abstracts, and body usage).

\subsection{Plural}

Plural uses are approx. 8 percent-points more likely to be \dialogue. This is significantly different than non-plural uses (Fisher's Exact $p \ll 0.05$).

\subsection{Proper Nouns}

We explore the spacy-identified Part of Speech (POS) tag for instances in abstracts, comparing PROPN to NOUNs. We find proper nouns to be approximately 10 percent-points less often \dialogue, a significant difference (Fisher's Exact $p \ll 0.05$). This result is counter intuitive under a hypothesis that \dialogue might be more formal. Instead it might be tied to later findings with word compounding where in longer terms, \dialog might be preferred. The difference, while statistically significant, is still fairly small.

\subsection{Compounds}

We hypothesize that when forming closed word compounds (which are when one concatenates words to form a new term without any spaces \cite{chicago2017}), the shorter \dialog will be preferred. We speculate such compounds might be especially prevalent in computing, as words are concatenated to make \texttt{IdentifierNames} in common programming languages.

Within the Dialog(ue) Publication titles and abstracts, we identify \NumWordcombos unique closed compounds. Of these, \NumDialogWordcombos use \dialog, or \DialogWordcomboFrac\% (C95 \DialogWordcomboFracCiLow-\DialogWordcomboFracCiHigh). This is significantly more than the overall \PercentPubDialog of total \dialog publications. This gives evidence in support of a hypothesis that in these longer terms the economy of \dialog is preferred.

Note, that word compounds are not explicitly included in the search for Dialogue Papers or in the ``Dialog(ue) Corpus'' (the search pattern for those includes word breaks). Thus this exploration is separate.

\section{Noun Phrase Analysis Details}\label{append:nounphrase}

\subsection{Sub-noun-chunks}
We break identified noun-chucks into sub-noun-chucks. For example, an occurrence of "online conversational \dialog{} system" is expanded to \{"online conversational \dialog{} system", "conversational \dialog{} system", "conversational \dialog{}", "\dialog{} system", "\dialog{}"\}.

\subsection{Filtering}

We want to ensure included sub-noun-phrases are used in sufficient quantities to enable analysis. For each sub-noun-phrase we find the set of authors using the term. We filter to sub-noun-phrases which are used by least 50 unique author names. To control for papers with a very large number of authors, we only select the first and last authors. This helps prevent including sub-noun-phrases used only by a small number of papers or small number of authors.

\subsection{Linear Modeling}\label{append:nounlinear}

We use the filtered phrases as exogenous variables in a logistic regression model. The variables of this model are not one-hot. Instead for a given noun phrase, multiple variables might be true for each sub-noun-phrase (e.g., the sub-phrase "spoken dialog(ue) system" would have three variables set for {"spoken dialog(ue) system", "spoken dialog(ue)", "dialog(ue)"} in addition to a constant variable). We model this using the \texttt{statsmodels} package \cite{seabold2010statsmodels} which provides p-values. Given the large number of variables, we have elevated risk of Type I errors. Thus we apply Benjamini-Hochberg correction \cite{Benjamini1995ControllingTF} which limits the Type I false discovery rate to our selected $\alpha$ of 0.05.

\begin{figure}[tbh]
    \centering
    \includegraphics[width=0.99\columnwidth]{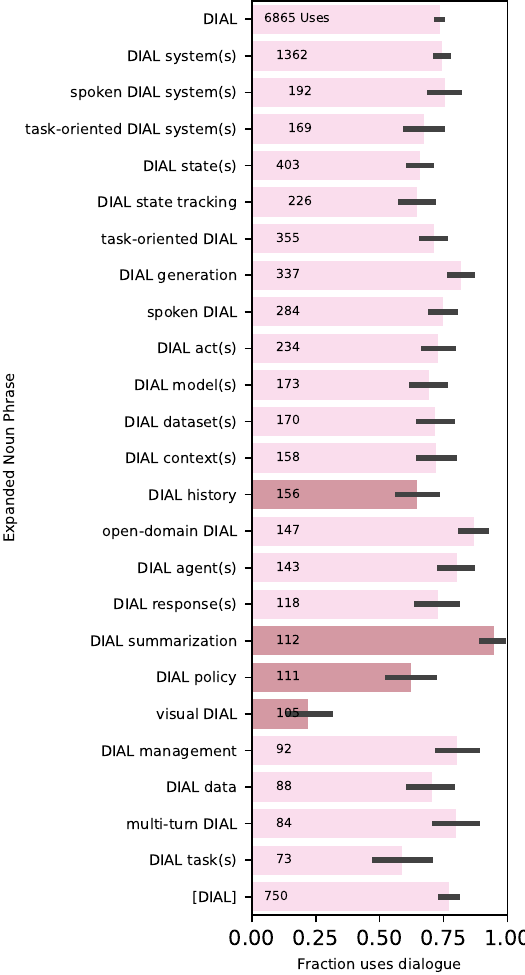}
    \caption{Top 25 sub-noun-phrases. Indentation of counts on the left side of each bar is used to emphasize how more complex sub-noun-phrases can be narrowing versions others. The [DIAL] category is a special category for uses of dialog(ue) which is independent and not part of a noun-phrase. Some quirks (such as dialog(ue) rarely being used a verb or with dependency parser failures) leads to not every example being represented here.}
    \label{fig:fullphrases}
\end{figure}

\section{Home/native nationality estimation}

We attempt to estimate where an author might have learned English spelling. To do this we search author names using the Brave Web Search API \footnote{https://brave.com/search/api/}, appending on the terms "institution computer science language". We then prompt GPT-4 to use these results to estimate the name's nationality informed by the web search results.

We find GPT-4 demonstrates impressive reasoning capabilities (for example, knowing the subtleties of common Turkish names, or identifying the location of small cities mentioned in the search results). However, it is often unable to identify to a correct location.

We ran this process for approx 800 authors. We then took the estimated nationality for the first author, getting estimates for approx 1300 papers. While countries like China had a large sample (over 300 authors), countries like US was labeled for less than 50 authors. Other categories like "unknown", where GPT-4 gave no answer was over 200 authors. This statistical power for detecting any difference small.
\begin{figure}
    \centering
    \includegraphics[width=0.99\linewidth]{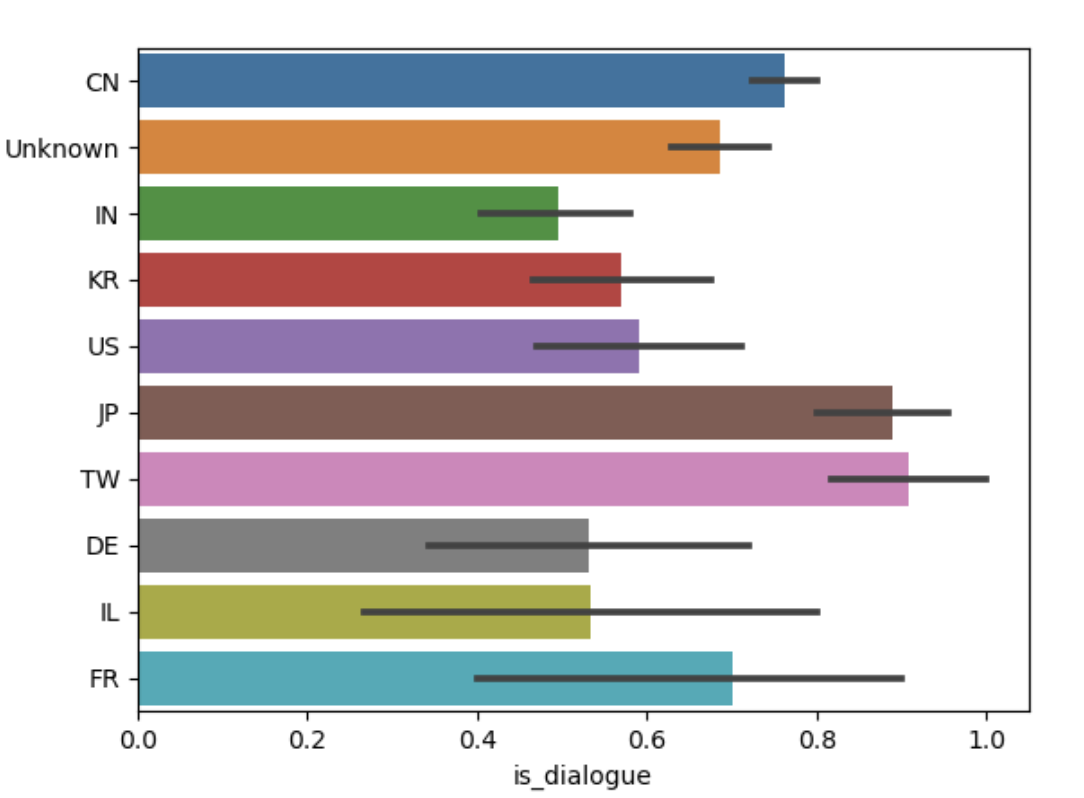}
    \caption{Data from estimated home nationality. Identification is to noisy to make clear conclusions for the amount of available data. The "Unknown" category might skew towards some nationalities in difficult to quantify ways. Thus we caution conclusion here.}
    \label{fig:homenationality}
\end{figure}
\label{append:nationality}

\section{Exploring Intentionality and Misspellings}\label{append:intentionality}
One possible explanation for the use of \dialog in Computing, that could also explain the elevated use in the fields Biology and Physics but not fields like History and Political Science, is along the lines of ``people in STEM are just bad at spelling''. We can sympathize with this reasoning. This "bad at English spelling" phenomenon might be independent of a field's level of non-native English speakers (as our findings in \autoref{natinfluence} suggest nationality is only a weak explainer). One method of exploring this deeper is looking at co-ocurance of \dialog and \dialogue with cases that are unambiguous misspellings.

Identifying misspellings is challenging in scientific work due to the use of jargon and rare terms. As a preliminary analysis of this, we prompt GPT-4-mini to extract misspellings in CS Dialogue Publication abstracts, followed by some heuristic filters to further process (for example, filtering some errors that were likely only the result of abstract parsing errors). This process estimates that approximately 10\% of CS Dialog(ue) Publications have likely spelling errors in their titles or abstracts (excluding any view about "dialog(ue)"). A Fisher Exact Test indicates there is not evidence there are differences between the rates among \dialog, \dialogue, and \both papers. Thus this gives some evidence that one category of authors might not be unintentionally misspelling at a higher rate.

More thorough analysis would expand to other fields to get more statistical power, or analyze the bodies of the papers. This is left as future work.

\section{LM Embedding Details}\label{append:lm}

Prior work has shown how Masked Language Models (MLMs) can be used for tasks of word sense induction \cite[\textit{inter alia}]{Wiedemann2019DoesBM, Amrami2019TowardsBS}. Inspired by this, we use MLMs to attempt to identify contextual senses that influence dialog(ue) spelling.

We select dialog(ue) uses and the surrounding context. We then mask out uses of dialog(ue)(s) using values from the RoBERTa tokenizer. We pass this through the RoBERTa cased model and extract the top layer embeddings of a given masked dialog(ue) use. This contextual embedding is a 768-dimensional vector. The RoBERTa model is specifically trained to predict the value of a masked token given context, thus we expect this vector to have predictive signal on use given context. We fit a logistic regression model on the extracted vectors with 10-fold cross-validation. The accuracy showed no improvement over the baseline of always predicting \dialogue (0.725 vs 0.739 baseline). We also tried other models like a multilayer perception with one hidden layer, k-means, and Gaussian mixture modeling, also all finding no improvement. This suggests limited influence of context on the spelling.

We did not explore fine-tuning the entire RoBERTa model. We note how, unlike other scenarios like adapting BERT/RoBERTa to tasks like sentiment classification or entailment, masked prediction was the original task of the model. Thus, there is possibly limited room for fine tuning. This is however a potential area of future work.

\section{Source Code Use}\label{sec:sourcecode}

We speculate that computer scientists might be influenced by code switching between natural language (NL) and programming languages (PL). As the PL writing is designed to be read by other humans, it can often have much in common with NL writing \cite{Allamanis2017ASO}, and might carry over.

We explore two corpora. For a broad view of source code, we sample 1 million python files from The Stack dataset \cite{Kocetkov2022TheStack}. We find 9543 files with dialog(ue) with 89\% using exclusively \dialog. Manually examining a subsample, we observe these are often for uses of GUI libraries, as suggested by \refexplan{exp:computing_specific}.

To narrow a bit more on NLP/AI, we use a dataset sourced from PyTorch-using repositories \cite{Xu_code-autocomplete_Code_AutoComplete}. Here we find 359 lines of code with dialog(ue) in approx 5.2M lines. \dialog represents 40\% of dialog(ue) lines, thus more than the ~20-25 percent-points than in paper's prose.\footnote{We must note that this sourecode dataset is at least a few years out of date.}

These elevated uses of \dialog might give partial evidence for some carryover into NLP/AI writing. We acknowledge though a more precise analysis might involve a process for identifying and searching for openly available source code for the research publications when available. This is left as future work.










\section{Additional Literature Review}

For direct comparison of \dialog and \dialogue we find we must mostly rely on web articles. \citet{LanguageTool2023} discusses the three explanations in \autoref{sec:intro}, and also mentions Google Ngrams data. \cite{writingexplained2024dialog} also discusses ngram data, and makes the strong claim that the only use for \dialog is for ``\dialog box''. \citet{august2012dialogue} briefly discusses both uses, and makes clearest arguments against \dialog, pointing out the relative rarity of ``monolog'' and ``epilog''.

Academic work such as \citet{shenguiding} discussed some historical use of computing terms. 

\citet{mcculloch_because_2019} extensively covered how technology and the internet has influenced language. As an example, she highlights aspects like computerized spell check influences on regularity, such as consistent British or American English in a document depending on the selected spell checker (p 46-47). While dialog(ue) is not discussed in the book, one might speculate that its permissive status in common spell checkers helped perpetuate a mixing.

\citet{venezky1999american} discusses American English spelling evolution. He discusses some of the role of American dictionaries (which seem uncommitted to \dialog \cite{MerriamWebster2024}), and also highlights the inconsistencies. \citet{sebba2007spelling} covers orthography more broadly, including its social influences. Given we find limited dialog(ue) variability by noun-phrase jargon or by experience level, it seems likely there is not social signaling occurring with dialog(ue) use.

We believe our work is the first to focus on dialog(ue) orthography, in particular from the important perspective of NLP/AI research that is now growing to reach millions or billions of users.

\end{document}